\documentclass[conference]{IEEEtran}
\IEEEoverridecommandlockouts
\usepackage{cite}
\usepackage{amsmath,amssymb,amsfonts}
\usepackage{algorithmic}
\usepackage{graphicx}
\usepackage{textcomp}
\usepackage{pifont}
\usepackage{xcolor}
\def\BibTeX{{\rm B\kern-.05em{\sc i\kern-.025em b}\kern-.08em
    T\kern-.1667em\lower.7ex\hbox{E}\kern-.125emX}}

\begin{document}

\bibliographystyle{IEEEtran}

\title{\textcolor{white}{aaaaaaaa} \\ 
Angle-I2P: Angle-Consistent-Aware Hierarchical Attention for Cross-Modality Outlier Rejection}

\author{Muyao Peng*, Shun Zou*, Pei An, You Yang and Qiong Liu

\thanks{Muyao Peng, Shun Zou, Pei An, You Yang, and Qiong Liu are with School of Electronic Information and Communications, Huazhong University of Science and Technology. (Email: \{muyao99, zoushun, anpei96, yangyou, q.liu\}@hust.edu.cn)}

\thanks{* The first two authors contributed equally.}

\thanks{Corresponding to yangyou@hust.edu.cn}
}

\maketitle

\begin{abstract}
Image-to-point-cloud registration (I2P) is a fundamental task in robotic applications such as manipulation, grasping, and localization. Existing deep learning-based I2P methods seek to align image and point cloud features in a learned representation space to establish correspondences, and have achieved promising results. However, when the inlier ratio of the initial matching pairs is low, conventional Perspective-n-Points (PnP) methods may struggle to achieve accurate results. To address this limitation, we propose Angle-I2P, an outlier rejection network that leverages angle-consistent geometric constraints and hierarchical attention. First, we design a scale-invariant, cross-modality geometric constraint based on angular consistency. This explicit geometric constraint guides the model in distinguishing inliers from outliers. Furthermore, we propose a global-to-local hierarchical attention mechanism that effectively filters out geometrically inconsistent matches under rigid transformation, thereby improving the \textit{Inlier Ratio} (IR) and \textit{Registration Recall} (RR). Experimental results demonstrate that our method achieves state-of-the-art performance on the 7Scenes, RGBD Scenes V2, and a self-collected dataset, with consistent improvements across all benchmarks.
\end{abstract}


\section{Introduction}
Image-to-Point Cloud Registration is a basic task in numerous downstream tasks like robot manipulation, simultaneous localization and mapping (SLAM) and autonomous driving \cite{wu2024economic, murai2025mast3r, an2024ol,an2025enhance}. It aims to calculate the alignment transformation $\mathbf{T}\in SE(3)$, which contains a rotation matrix $\mathbf{R} \in SO(3)$ and a three-dimension translation vector $\mathbf{t} \in \mathbb{R}^{3}$. Existing methods have achieved promising results in image-to-point cloud registration tasks\cite{wang2021p2, feng20192d3d, an2024ol, bie2025graphi2p}. However, when training with limited data or in the presence of significant point cloud noise, numerous incorrect pixel-to-point cloud correspondences may arise. It will significantly affect the performance of Perspective-n-Points (PnP) \cite{peng2025ldfi2p}. 

In order to reject outliers from initial correspondences, Gomatch \cite{zhou2022geometry} proposes a linear-layer-based classifier module. It takes concatenated geometric features from putative 2D-3D keypoint pairs to filter out unreliable matches (via a configurable threshold parameter). GraphI2P \cite{bie2025graphi2p} and its preceding work \cite{bie2024build, bie2024image} employ monocular depth estimation to predict depth from an image and subsequently back-project the image into 3D space. This converts the image-to-point cloud registration problem into a point cloud registration (PCR) problem. Then they construct an adjacent matrix and apply a graph neural network (GNN) to enhance pattern consistency among correct correspondences. A key drawback is that the inherent scale ambiguity in depth estimation compromises performance. Consequently, the key challenges in eliminating incorrect image-to-point cloud matches are: 1) how to minimize the modality gap between images and point clouds as much as possible, and 2) how to design effective geometric constraints to filter out non-conforming outliers and enhance registration performance.

\begin{figure}
    \centering
    \includegraphics[width=1.0\linewidth]{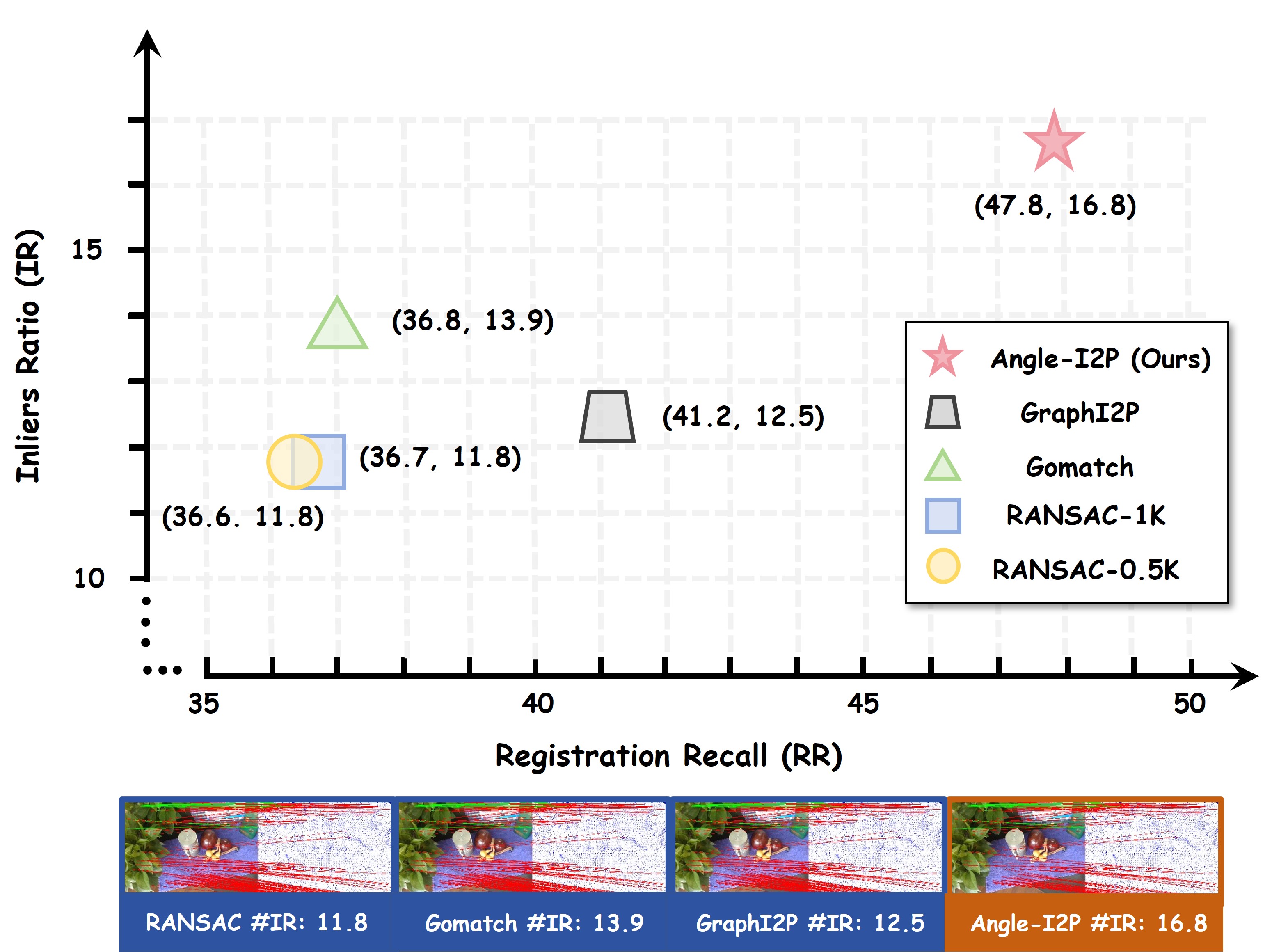}
    \caption{Registration performance of existing image-to-point cloud outliers rejection methods on \textit{Self-collected Datasets}. Existing works \cite{fischler1981random, zhou2022geometry, bie2025graphi2p} have made efforts to reject outliers in I2P tasks. In this paper, we propose Angle-I2P. Through back-project, scale alignment and an angle-consistent-aware hierarchical attention, our method (\textcolor{red}{\ding{72}}) demonstrates enhanced potential for practical deployment in real-world scenarios.}
    \label{Figure 1}
\end{figure}

To solve the above problems, we propose Angle-I2P. It has better potential for practical deployment, shown as Fig. \ref{Figure 1}. To address the problem of mismatches in cross-modality registration, we propose a robust inlier criterion by transforming the image-point cloud matching problem into 3D space via monocular depth estimation and aligning scales globally. This method effectively leverages noise-robust 3D geometric properties to achieve efficient outlier rejection. To mitigate the influence of scale noise on inlier discrimination, we propose a novel module named \textit{Angle-Consistent-Aware Hierarchical Attention}. It effectively integrates angle-consistent features from global to local contexts. This integration enhances the capability to discriminate outliers, which in turn improves the overall registration performance. Extensive experiments on three challenging benchmarks demonstrate clear superiority of our method. 

\section{Related Works}

\subsection{Image-to-Point Cloud Registration}

Existing works can be divided into two categories: detect-then-match based methods and no-detector based methods. The first methods mostly detect the keypoints in 2D and 3D space seperately and match them \cite{feng20192d3d, pham2020lcd, wang2021p2, zhou2022geometry}. MinCD-PnP \cite{an2025mincd} simplifies blind PnP into a Chamfer distance minimization task between 2D and 3D keypoints. It proposes MinCD-Net, a lightweight multi-task module that improves robustness to noise and outliers in image-to-point-cloud registration. This kind of methods lack interaction between two modality, result in low registration performance. No-detector based methods aim to establish cross-modality correspondences to support pose estimation by rationally fusing modalities, thereby enabling the expression and alignment of cross-modality features within a unified latent space. 2D3D-MATR \cite{li20232d3d} proposed a attention based methods to fuse the feature of different modality. FreeReg \cite{wang2023freereg} and DiffReg \cite{wu2024diff} leverage diffusion models by formulating correspondence estimation as a denoising diffusion process within the doubly stochastic matrix space, where cross-modality features (image and point cloud) are iteratively refined through a lightweight transformer during reverse sampling to robustly establish geometrically consistent matches. GraphI2P \cite{bie2025graphi2p} and its predecessor works \cite{bie2024build, bie2024image} transfer the modality gap to the distribution gap by introducing a monocular depth estimator to convert the images to point cloud.  

\subsection{Graph-Based Outliers Rejection Methods}
Outliers Rejection has been widely studied in point cloud registration (PCR) \cite{fischler1981random,yang2020teaser,barath2018graph,zhang2024fastmac,zhang2025mac++,zhang20233d,zhang2025hypergct}. RANSAC \cite{fischler1981random} is the most popular method in this field, subsequent work achieves more robust results by incorporating graph structures \cite{barath2018graph, barath2021graph, zhang2024fastmac, zhang2025mac++}. However, traditional methods are often constrained by limitations such as computational efficiency. In contrast, learning-based approaches can effectively approximate the optimal solution \cite{zhang20233d, zhang2025muscle, wang20243dpcp}. As our baseline, GraphSCNet \cite{qin2023deep} proposes the first learning-based outlier rejection method for non-rigid point cloud registration. It presents a graph-based local spatial consistency measure that leverages the local rigidity of deformations, combined with an attention-based correspondence embedding module to enhance feature representation and accurately filter outliers. HyperGCT \cite{zhang2025hypergct} proposes a dynamic hypergraph neural network-based method that learns high-order geometric constraints for robust 3D point cloud registration, achieving state-of-the-art performance and improved generalization across diverse datasets. To the best of our knowledge, existing works don't focus on cross-modality outliers rejection.

\section{Method}

\begin{figure*}
    \centering
    \includegraphics[width=0.9\linewidth]{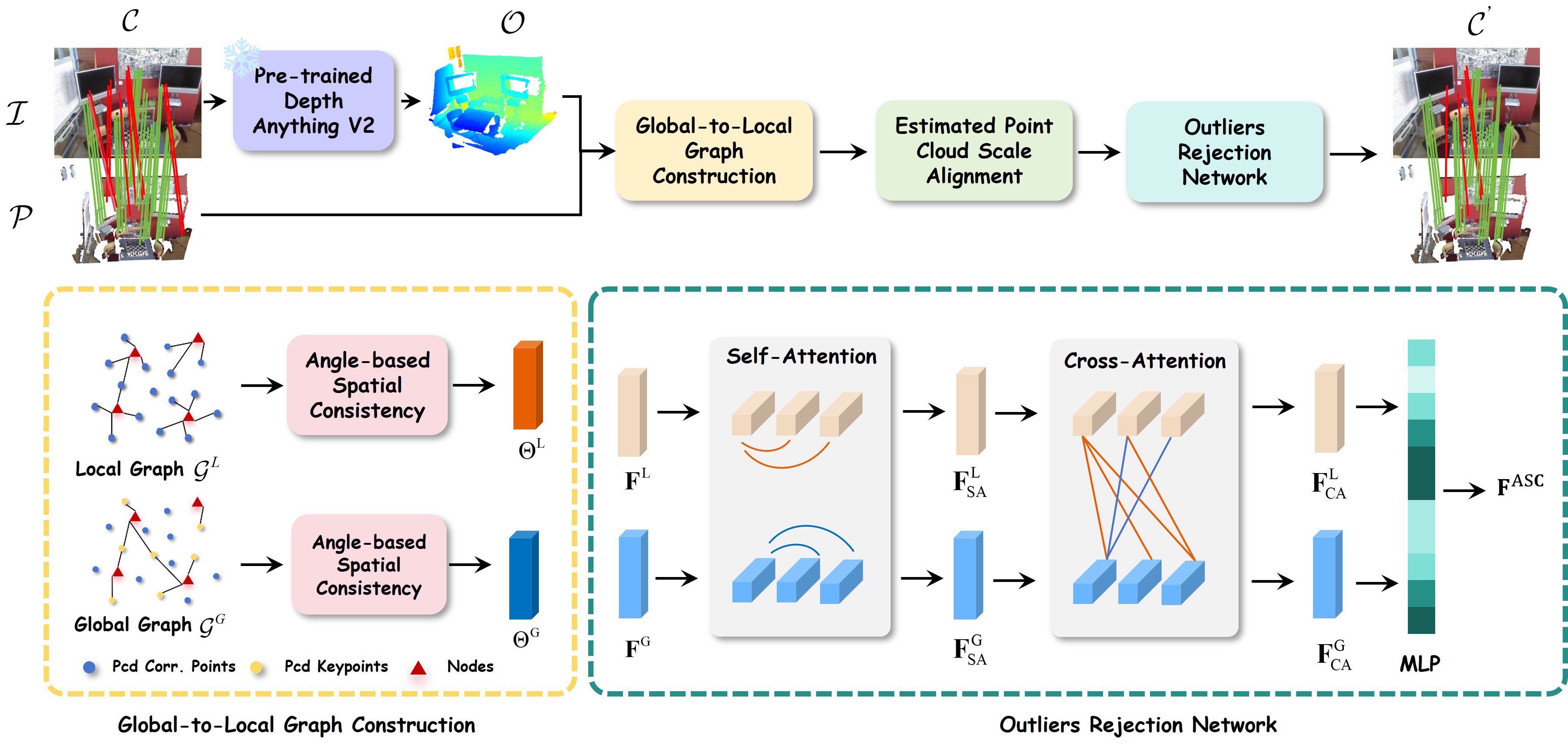}
    \caption{Pipeline of the Angle-I2P. First, we back-project the image $\mathcal{I}$ to point cloud $\mathcal{O}$ using the depth estimated from pre-trained monocular depth estimator. Then we construct global-wise and local-wise graph $\mathcal{G}^{\mathrm{G}}$ and $\mathcal{G}^{\mathrm{L}}$. We use them to compute the global-wise and local-wise angle-based spatial consistency. Also, we use them to get the feature $\mathbf{F}^{\mathrm{G}}$ and $\mathbf{F}^{\mathrm{L}}$. Then $\mathbf{F}^{\mathrm{G}}$ and $\mathbf{F}^{\mathrm{L}}$ are put into the outliers rejection network to obtain a more robust correspondences $\mathcal{C}'$}
    \label{Fig. 2: Pipeline of the proposed method.}
\end{figure*}

\subsection{Overview}\label{section 3.1 Overview}

Given an image $\mathcal{I}\in\mathbb{R}^{\mathrm{H}\times\mathrm{W}\times3}$ and a point cloud $\mathcal{P}\in\mathbb{R}^{\mathrm{N}\times3}$, existing image-to-point cloud registration networks \cite{wang2021p2, li20232d3d, wang2023freereg} aim to get the correspondences $\mathcal{C}=\{\mathbf{c}_{i}=(\mathbf{p}_{i},\mathbf{q}_{i})\in\mathbb{R}^{5}|~\mathbf{p}_{i}\in\mathcal{I},\mathbf{q}_{i}\in\mathcal{P}\}^{N}_{i=1}$ first. Then $\mathcal{C}$ is used to calculate the alignment transformation $\mathbf{T}\in SE(3)$, which contains a rotation matrix $\mathbf{R} \in SO(3)$ and a three-dimension translation vector $\mathbf{t} \in \mathbb{R}^{3}$. 

Despite achieving satisfactory results, existing image-point cloud registration methods \cite{li2021deepi2p, li20232d3d} still underperform when trained with limited data and validated across different scenes. Therefore, it is necessary to filter outliers from the initial set of correspondences $\mathcal{C}$ to obtain a more robust correspondence set $\mathcal{C'}$. 

In this paper, an outliers rejection network is present for I2P tasks. First, we convert images into point clouds using a monocular depth estimator, and then enforce global-to-local spatial consistency to constrain the cross-modality data. Subsequently, to mitigate the impact of scale errors on spatial consistency computation, we designed an angular metric-based approach for calculating spatial consistency (Sec. \ref{section 3.2 Global-to-Local Spatial Consistency}). Finally, we estimated and aligned the scale of the estimated point cloud with the ground truth. After rescale the estimated points cloud, we tailor a global-to-local cross-attention to constrained the geometric structure of correspondences from the global to local level.  Thereby we believe it can effectively filter out image-point cloud correspondences under rigid transformations (Section \ref{Section 3.3 Outliers Rejection Network}). The pipeline of our method is shown as Fig. \ref{Fig. 2: Pipeline of the proposed method.}

\subsection{Angle-Based Spatial Consistency}
\label{section 3.2 Global-to-Local Spatial Consistency}
\noindent \textbf{Estimated Point Cloud Generation.} In order to solve the modality gap between images and point clouds, we convert the images to point clouds using depth estimation follow GraphI2P \cite{bie2025graphi2p}. First, we use Depth Anything V2 \cite{yang2024depth} to estimate the depth $\mathcal{D}$ of the image $\mathcal{I}$, expressed as (\ref{eq1}). Then we use the intrinsics of the camera to convert image to point cloud, expressed as (\ref{eq2}).

\begin{equation}
    \label{eq1}
    \mathcal{D}=\mathrm{DepthAnythingV2}(\mathcal{I})
\end{equation} 

\begin{equation}
    \label{eq2}
    \mathcal{O}^{\mathrm{gt}}=\pi^{-1}(\mathcal{D}^{\mathrm{gt}}, \mathcal{K})=\pi^{-1}(s\mathcal{D}+t, \mathcal{K})=s\mathcal{O}+bias
\end{equation}

\noindent in which $\mathcal{O}=\{\mathbf{o}_{i}\}^{N}_{i=1}\in\mathbb{R}^{\mathrm{N}\times3}$ is the estimated point cloud and $\mathcal{O}^{\mathrm{gt}}=\{\mathbf{o}^{\mathrm{gt}}_{i}\}^{N}_{i=1}\in\mathbb{R}^{\mathrm{N}\times3}$ is the point cloud computed using the ground-truth depth. $\pi^{-1}(\cdot)$ is the back projection method and $\mathcal{K}$ is the intrinsics of the camera. $s,t$ are the scale and bias between the ground-truth depth and the estimated depth. $bias$ is the bias between the estimated point cloud and the ground-truth point cloud. Then we could get the estimated correspondences $\mathcal{C}^{\mathrm{est}}=\{(\mathbf{o }_{i},\mathbf{q}_{i})\in\mathbb{R}^{6}|~\mathbf{o}_{i}\in\mathcal{O},\mathbf{q}_{i}\in\mathcal{P}\}^{N}_{i=1}$.

\noindent \textbf{Angle-Based Spatial Consistency.} Any two correspondences in $\mathcal{C}$ should be satisfied with (\ref{eq3}). 

\begin{equation}
    \label{eq3}
    \Vert \mathbf{q}_{i}-\mathbf{q}_{j}\Vert \approx \Vert \pi^{-1}(\mathbf{p}_{i}, d^{\mathrm{gt}}_{i},\mathcal{K})-\pi^{-1}(\mathbf{p}_{j}, d^{\mathrm{gt}}_{j},\mathcal{K}) \Vert
\end{equation}

\noindent in which $d^{\mathrm{gt}}_{i}$ is the ground-truth depth value of pixel $\mathbf{p}_{i}$. But due to the scale ambigious cased by the monocular depth estimate function, (\ref{eq3}) is converted to (\ref{eq4}):

\begin{equation}
    \label{eq4}
    \begin{split}
        &~~~~\Vert \mathbf{q}_{i}-\mathbf{q}_{j}\Vert \\
        &\approx \Vert s *\pi^{-1}(\mathbf{p}_{i}, d_{i},\mathcal{K}) - s * \pi^{-1}(\mathbf{p}_{j}, d_{j},\mathcal{K})\Vert \\
        &\approx s * \Vert \pi^{-1}(\mathbf{p}_{i}, d_{i},\mathcal{K}) - \pi^{-1}(\mathbf{p}_{j}, d_{j},\mathcal{K})\Vert
    \end{split}
\end{equation}

\noindent in which $d_{i}$ is the estimated depth value of pixel $\mathbf{p}_{i}$. Due to the presence of scale errors, the commonly used spatial consistency constraint in point cloud outliers rejection networks \cite{zhang2025mac++, zhang2024fastmac, qin2023deep} is relaxed, leading to mismatches. Therefore, we propose a scale-invariant spatial consistency calculation method to constrain the geometric relationships between matching pairs. First, the centroid of each point cloud, $\mathbf{o}_{c}$ and $\mathbf{p}_{c}$, was computed, expressed as (\ref{eq5}).

\begin{equation}
    \label{eq5}
    \mathbf{o}_{c}=\frac{1}{N}\sum_{i=1}^{N}\mathbf{o}_{i},~~\mathbf{p}_{c}=\frac{1}{N}\sum_{i=1}^{N}\mathbf{p}_{i}
\end{equation}

\noindent where $N=|\mathcal{C}|$ is the number of the correspondences.

Next, all points were translated to a coordinate system centered at their respective centroid to eliminate translational degrees of freedom, denoted as $\hat{\mathbf{o}}_{i}$ and $\hat{\mathbf{p}}_{i}$. It can also be interpreted as a vector pointing from the origin toward the point cloud. For each pair of matching points, we compute the cosine of the angle between their corresponding vectors and subtract these values, as detailed in (\ref{eq6}).  

\begin{equation}
    \label{eq6}
    \mathrm{cos}(\alpha^{\mathcal{I}}_{ij})=\frac{\hat{\mathbf{o}}_{i} \cdot \hat{\mathbf{o}}_{j}}{\Vert \hat{\mathbf{o}}_{i} \Vert \Vert \hat{\mathbf{o}}_{j} \Vert}
    =\frac{s^{2}~(\hat{\mathbf{o}}^{\mathrm{gt}}_{i} \cdot \hat{\mathbf{o}}^{\mathrm{gt}}_{j})}{s^{2}~\Vert \hat{\mathbf{o}}^{\mathrm{gt}}_{i} \Vert \Vert \hat{\mathbf{o}}^{\mathrm{gt}}_{j} \Vert} =\frac{\hat{\mathbf{o}}^{\mathrm{gt}}_{i} \cdot \hat{\mathbf{o}}^{\mathrm{gt}}_{j}}{\Vert \hat{\mathbf{o}}^{\mathrm{gt}}_{i} \Vert \Vert \hat{\mathbf{o}}^{\mathrm{gt}}_{j} \Vert}
\end{equation}

\noindent in which $\alpha^{\mathcal{I}}_{ij}$ is the angle between the two vectors. Through derivation, we have proven that (\ref{eq6}) is independent of scale $s$. Therefore, we can compute spatial consistency using the estimated point cloud, expressed as (\ref{eq7}):

\begin{equation}
    \label{eq7}
    \theta_{ij}=[1-\delta^{2}_{ij}/\sigma^{2}_{d}]_{+}
\end{equation}

Here, $[\cdot]_{+}=\max(0,\cdot)$, $\delta_{ij}=\big \vert \vert \cos(\alpha^{\mathcal{I}}_{ij}) \vert -  \vert \cos(\alpha^{\mathcal{P}}_{ij}) \vert \big \vert$ represents the difference in the cosine values of the vector angles between the two point clouds, and $\sigma_{d}$ is a hyper-parameter controlling sensitivity to distance variation. If both points are inliers, $\delta_{ij}$ should be small, causing $\theta_{ij}$ to approach 1. Conversely, if at least one point is an outlier, $\delta_{ij}$ tends to be large and therefore $\theta_{ij}$ should be 0. This provides strong geometric justification for rejecting outliers in rigid scenarios, and we argue that it imposes equivalent constraints to traditional spatial consistency.

\subsection{Outliers Rejection Network}
\label{Section 3.3 Outliers Rejection Network}

\begin{figure*}
    \centering
    \includegraphics[width=1.0\linewidth]{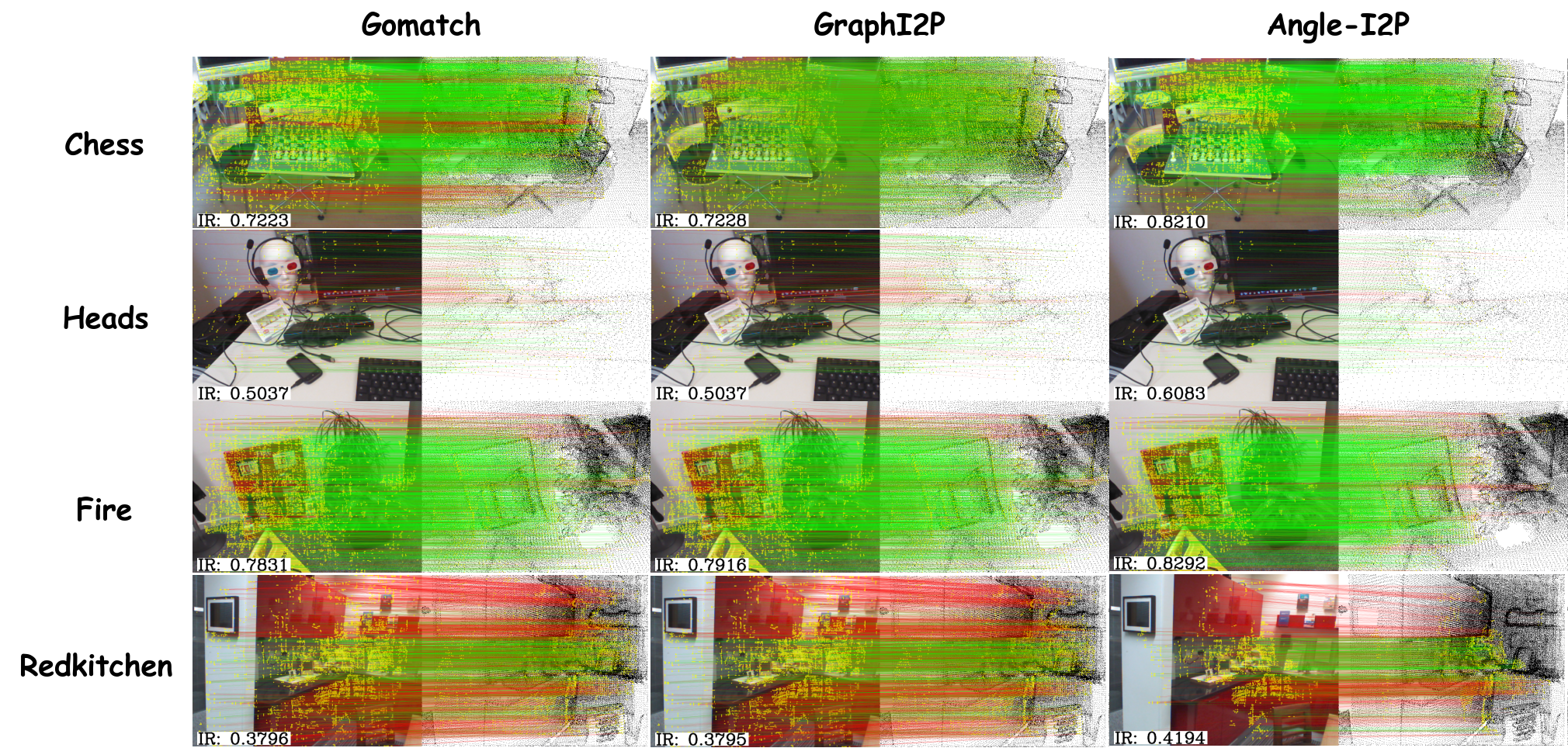}
    \caption{Visualization of the outliers rejection results of each method. We show four selected scenes in 7Scenes datasets. \textcolor{green}{Green lines} represent inliers. \textcolor{red}{Red lines} represent outliers. Our method achieve the best performance of all methods (Compare the 3rd column to the 1st and 2nd column).}
    \label{Figure 2: Visualization of 7Scenes Datasets.}
\end{figure*}

We follow GraphSCNet \cite{qin2023deep} to use coordinates of points as initial features. For scale discrepancies exist between the estimated and actual point clouds, it will introduces biases into the extracted features, which ultimately degrades the model's performance. So we roughly estimated a global average scale factor between the reconstructed point cloud $\mathcal{O}$ and the point cloud $\mathcal{P}$. Since rotation preserves the Euclidean distance between points, the distance from each point to its cloud's centroid is calculated, which are recorded as $\mathbf{l}^{\mathcal{I}}_{i}$ and $\mathbf{l}^{\mathcal{P}}_{i}$. Finally, the scale factor $s$ between the two point clouds is estimated using the ratio of corresponding distances: 

\begin{equation}
    \label{eq8}
    s^{\mathrm{est}}=\frac{1}{N}\sum_{i=1}^{N}\frac{\mathbf{l}^{\mathcal{I}}_{i}}{\mathbf{l}^{\mathcal{P}}_{i}}=\frac{1}{N}\sum_{i=1}^{N}\frac{\Vert\mathbf{o}_{i}-\mathbf{o}_{c}\Vert}{\Vert\mathbf{p}_{i}-\mathbf{p}_{c}\Vert}
\end{equation}

We rescale the estimated point cloud $\hat{\mathbf{o}}_{i}$ to $\tilde{\mathbf{o}}_{i}=s^{\mathrm{est}}\times\hat{\mathbf{o}}_{i}$ using the estimated scale $s^{\mathrm{est}}$. Using the estimated point cloud coordinates as features may introduce initial errors. To address this, we further incorporate the point cloud normals as constraints. Then we could get the initial feature $\mathbf{f}_{i}$: 

\begin{equation}
    \label{eq9}
    \mathbf{f}_{i}=[\tilde{\mathbf{c}}_{i};\sin(2^{-1}\tilde{\mathbf{c}}_{i});\cos(2^{-1}\tilde{\mathbf{c}}_{i});\mathbf{n}^{\mathcal{I}}_{i};\mathbf{n}^{\mathcal{P}}_{i}] \in \mathbb{R}^{24}
\end{equation}

\noindent in which $\tilde{\mathbf{c}}_{i}$ is the conbination of $\tilde{\mathbf{o}}_{i}$ and $\hat{\mathbf{p}}_{i}$. $\mathbf{n}^{\mathcal{I}}_{i}$ and $\mathbf{n}^{\mathcal{P}}_{i}$ are the normals of the point cloud and estimated point cloud. The initial feature is then put into a linear layer block to get the $\mathbf{F}^{\mathrm{init}}$.

\noindent \textbf{Angle-Consistent-Aware Hierarchical Attention.} For local spatial consistency, we sample a set of nodes $\mathcal{V}=\{\mathbf{v}_{j}\in\mathbb{R}^{3}|~j=1,\cdots,V\}$ from the original point cloud $\mathcal{P}$. And then get a set of correspondences assigned to a node $\mathbf{v}_{j}$, denote as $\mathcal{C}_{j}=\{\mathbf{c}_{i}\vert~\mathbf{c}_{i}\in\mathcal{N}_{j}\}^{K}_{i=1}$, in which $\mathcal{N}_{j}$ is the k-nearest neighbours of $\mathbf{v}_{j}$.  

For global spatial consistency, we employ the method from P2-Net\cite{wang2021p2} to select the top $M$ value-maximizing keypoints. These $M$ keypoints effectively represent the global geometric information of the scene or object. Subsequently, we assign each of these $M$ points to the nodes $\mathcal{V}$ using k-nearest neighbors approach, thereby constructing the global graph $\mathcal{G}^{\mathrm{G}}\in\mathbb{R}^{V \times K \times 3}$. 

Finally, global and local spatial consistency is computed by applying (\ref{eq7}) to $\mathcal{G}^{\mathrm{G}}$ and $\mathcal{G}^{\mathrm{L}}$. We iteratively use self-attention and cross-attention to refine the features. In self-attention, $\mathbf{F}^{\mathrm{Q}}$, $\mathbf{F}^{\mathrm{K}}$ and $\mathbf{F}^{\mathrm{V}}$ are local features or global features. In the cross-attention, we use either global or local features as the \textit{Query}, and the other type of features as the \textit{Key} and \textit{Value}. This integrates both global and local geometric information.

\begin{equation}
    \label{eq10}
    \mathbf{Q}=\mathbf{W}^{\mathrm{Q}}\mathbf{F}^{\mathrm{Q}}~~
    \mathbf{K}=\mathbf{W}^{\mathrm{K}}\mathbf{F}^{\mathrm{K}}~~
    \mathbf{V}=\mathbf{W}^{\mathrm{V}}\mathbf{F}^{\mathrm{V}}
\end{equation}

\noindent where $\mathbf{W}^{\mathrm{Q}}$, $\mathbf{W}^{\mathrm{K}}$ and $\mathbf{W}^{\mathrm{V}}$ is the projection weight. Inspired by GraphSCNet \cite{qin2023deep}, we reweight the attention score using the global angle-based spatial consistency and local angle-based spatial consistency.

\begin{equation}
    \mathrm{Attention}=\mathrm{Softmax}(\Theta\frac{\mathbf{Q}\mathbf{K}^{\mathrm{T}}}{\sqrt{d}})\mathbf{V}
\end{equation}

We expect this strategy to effectively enable feature fusion leveraging both global and local geometric information, thereby filtering out geometrically inconsistent outliers.

\noindent \textbf{Classification Head.} We put the angle-based spatial consistency aware feature $\mathbf{F}^{\mathrm{ASC}}$ into a MLP block to get the confidence score $\mathrm{score}$ of the inliers/outliers. At last, we could obtain a more robust correspondence set $\mathcal{C}^{'}=\{\mathbf{c}_{i}~|~\mathrm{score}_{i} \ge \tau \}^{L}_{i=1}$. 

\section{Experiments}

\subsection{Experimental Setup}

\begin{table}[t]
    \centering
    \caption{Evaluation Results on 7Scenes Datasets. \textbf{Boldfaced} numbers highlight the best and the second best are \underline{underlined}. $\uparrow$ means higher is better and $\downarrow$ means lower is better. $\dagger$ indicates that we independently reviewed the relevant code using the methodology described in the paper.}
    \resizebox{0.9\linewidth}{!}{%
    \begin{tabular}{l|cccccc}
    \hline    
    Model &IR$\uparrow$  &MRE$\downarrow$  &MTE$\downarrow$  &RR (@0.1m)$\uparrow$  \\ \hline
    RANSAC-0.5k \cite{fischler1981random} &44.6  &3.410  &0.081  &75.2  \\
    RANSAC-1k \cite{fischler1981random} &44.6  &\underline{3.093}  &\textbf{0.075}  &75.5   \\
    GoMatch \cite{zhou2022geometry} &45.0  &3.119  &\underline{0.076}  &\underline{76.2}  \\
    GraphI2P$^{\dagger}$ \cite{bie2025graphi2p} &\underline{45.3}  &\textbf{3.090}  &\textbf{0.075}  &76.0  \\
    Angle-I2P (Ours) &\textbf{49.5}  &3.237  &\textbf{0.075}  &\textbf{78.5}  \\ \hline
    \end{tabular}
    }
    \label{Table 1: Evaluation Results on 7Scenes Datasets}
\end{table}

\noindent \textbf{Implemental Details.} We conduct our experiments on a single NVIDIA GeForce RTX 4090 GPU and remain most settings in GraphSCNet \cite{qin2023deep}. Learning rate is set to $1\times10^{-4}$ and the weight decay is set to $1\times10^{-6}$. The $M$ and the number of neighbors $K$ in Section \ref{Section 3.3 Outliers Rejection Network} is set to 100 and 32.

\noindent \textbf{Baselines.} As there are no existing end-to-end image-to-point cloud outliers rejection network, we have selected and reproduced networks with outlier removal modules. We mainly compare with three methods: (1) RANSAC: a method which commonly used to reject outliers in PCR and I2P tasks. (2) GoMatch \cite{zhou2022geometry}, a method for inlier-outlier classification using a linear classification network. (3) GraphI2P \cite{bie2025graphi2p}, a method for inlier-outlier classification using a graph convolution network. Unfortunately, authors of GraphI2P don't public their code, we implement it based on the methodology described in their paper. It should be noted that we use Depth Anything V2 to get the virtual points. All initial correspondences $\mathcal{C}$ are obtained based on 2D3D-MATR \cite{li20232d3d}.

 \textbf{Metrics.} We evaluate our method with four metrics: (1) \textit{Inlier Ratio} (IR), the ratio of pixel-point correspondences whose 3D distance is below a certain threshold (\textit{i.e.}, 5cm). (2) \textit{Mean Rotation Error} (MRE): The mean value of the rotation error. (3) \textit{Mean Translation Error} (MTE): The mean value of the translation error. (4) \textit{Registration Recall} (RR): the ratio of corresponding 3D points whose reprojection distances (using the ground-truth pose and the estimated pose) exceed a given threshold (\textit{i.e.}, 0.1m).

\subsection{Evaluation Results on 7Scenes Datasets}

\begin{table}[t]
    \centering
    \caption{Evaluation Results on RGBD Scenes V2 Datasets. \textbf{Boldfaced} numbers highlight the best and the second best are \underline{underlined}. $\uparrow$ means higher is better and $\downarrow$ means lower is better. $\dagger$ indicates that we independently reviewed the relevant code using the methodology described in the paper.}
    \resizebox{0.9\linewidth}{!}{%
    \begin{tabular}{l|cccccc}
    \hline    
    Model &IR$\uparrow$  &MRE$\downarrow$  &MTE$\downarrow$  &RR (@0.1m)$\uparrow$  \\ \hline
    RANSAC-0.5k \cite{fischler1981random} &33.0  &2.052  &0.046  &43.1  \\
    RANSAC-1k \cite{fischler1981random} &33.0  &\underline{1.967}  &\underline{0.045}  &43.7   \\
    GoMatch \cite{zhou2022geometry} &33.0  &\underline{1.967}  &\underline{0.045}  &43.7  \\
    GraphI2P$^{\dagger}$ \cite{bie2025graphi2p} &\underline{33.2}  &\underline{1.967}  &\underline{0.045}  &\underline{43.8}  \\
    Angle-I2P (Ours) &\textbf{39.1}  &\textbf{1.887}  &\textbf{0.044}  &\textbf{44.7}  \\ \hline
    \end{tabular}
    }
    \label{Table 2: Evaluation Results on Cross-scene Registration}
\end{table}

\begin{table}[t]
    \centering
    \caption{Evaluation Results on Self-Collected Datasets. \textbf{Boldfaced} numbers highlight the best and the second best are \underline{underlined}. $\uparrow$ means higher is better and $\downarrow$ means lower is better. $\dagger$ indicates that we independently reviewed the relevant code using the methodology described in the paper.}
    \resizebox{0.9\linewidth}{!}{%
    \centering
    \begin{tabular}{l|cccccc}
    \hline    
    Model &IR$\uparrow$  &MRE$\downarrow$  &MTE$\downarrow$ &RR (@0.2m)$\uparrow$  \\ \hline
    RANSAC-0.5k \cite{fischler1981random} &11.8  &21.283  &0.330  &36.6  \\
    RANSAC-1k \cite{fischler1981random} &11.8  &18.316  &0.298  &36.7 \\
    GoMatch \cite{zhou2022geometry} &\underline{13.9}  &19.390  &\textbf{0.268}  &36.8  \\
    GraphI2P$^{\dagger}$ \cite{bie2025graphi2p} &12.5  &\underline{17.394}  &0.304  &\underline{41.2}  \\
    Angle-I2P (Ours) &\textbf{16.8}  &\textbf{16.035}  &\underline{0.282}  &\textbf{47.8}  \\ \hline
    \end{tabular}
    }
    \label{Table 3: Evaluation Results on Self-collected Datasets.}
\end{table}

\noindent \textbf{Datasets.} 7Scenes \cite{glocker2013real} is a datasets which contains seven different indoor scenes. We trained 2D3D-MATR \cite{li20232d3d} using full training set and then evaluated on both the full training set and the test set to generate initial correspondences. Consequently, we obtain matching results for 4,048 training pairs and 2,034 test pairs. For our experiments, the test set remained consistent with the validation set.

\begin{figure*}
    \centering
    \includegraphics[width=1.0\linewidth]{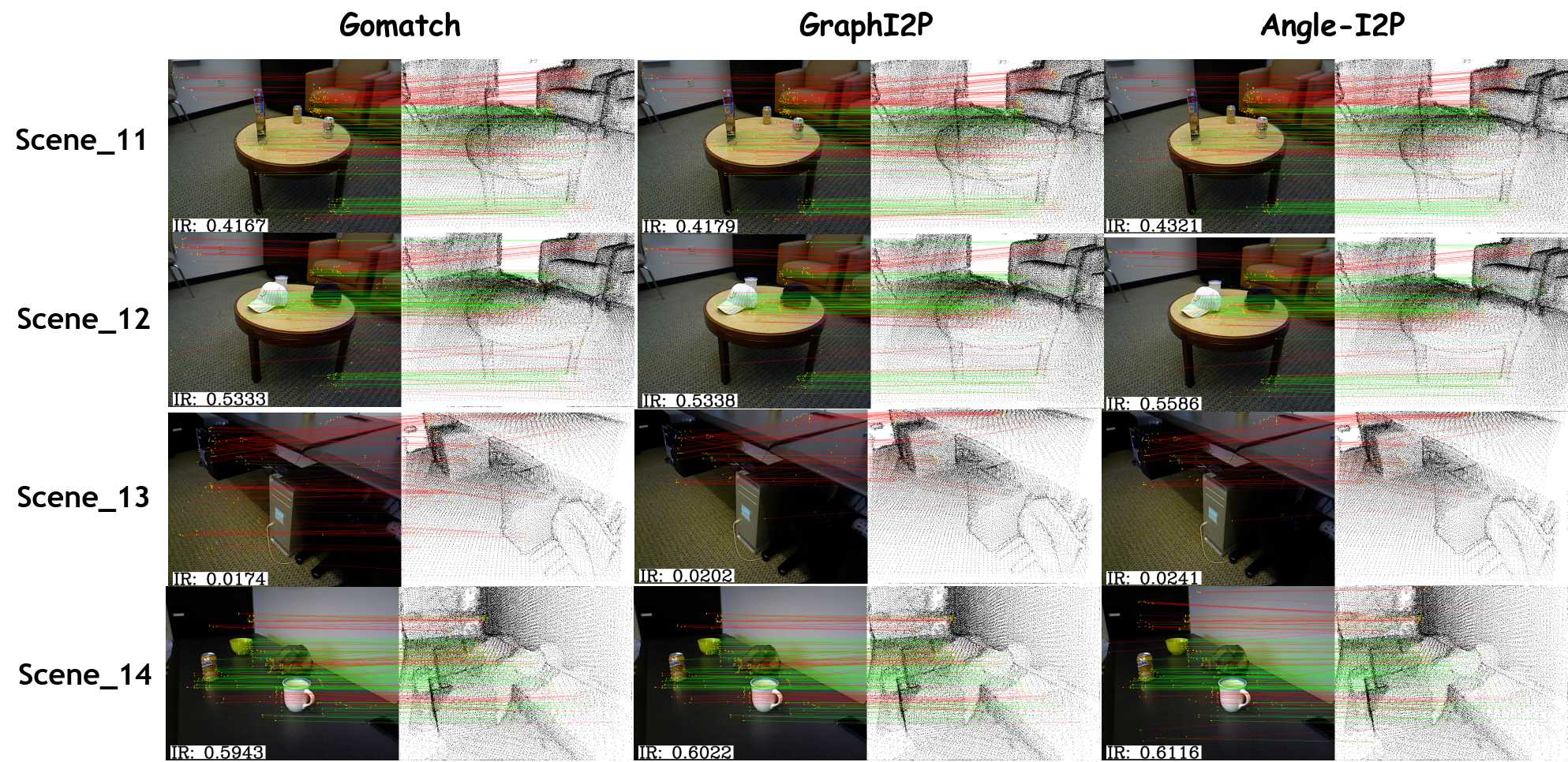}
    \caption{Visualization of the outliers rejection results of each method. We show the results of RGBDScenesV2 datasets. \textcolor{green}{Green lines} represent inliers. \textcolor{red}{Red lines} represent outliers. Our method achieve the best performance of all methods (Compare the 3rd column to the 1st and 2nd column).}
    \label{Figure 4: Visualization of RGBDScenesV2 Datasets.}
\end{figure*}

\noindent \textbf{Evaluation Results.} Since the test set shares the same scenes as the training set, we obtain more image-point cloud correspondences. The results are shown in Tab. \ref{Table 1: Evaluation Results on 7Scenes Datasets}. Our method outperforms the second best (GraphI2P  \cite{bie2025graphi2p}) by 4.2 percent on \textit{Inlier Ratio} and outperforms the second best Gomatch (Gomatch  \cite{zhou2022geometry}) 2.3 percent on \textit{Registration Recall}. Furthermore, our model demonstrates significantly improved localization performance. When utilizing filtered matches for pose estimation, it achieves higher localization accuracy. 

In order to better show our results, we visualize our results in Fig. \ref{Figure 2: Visualization of 7Scenes Datasets.} (The 3rd column is our method). It can be seen that our method can filter out more outliers than other methods.

\subsection{Evaluation Results on RGBDScenesV2 Datasets}
\label{Section 4.2 Evaluation Results on Cross-scene Registration.}

\begin{table*}[t]
    \caption{Ablation studies on 7Scenes datasets. \textbf{Boldfaced} numbers the best. Additionally, we present quantitative performance changes resulting from the removal of relevant modules. $\uparrow$ means higher is better and $\downarrow$ means lower is better.} 
    \centering
    \begin{tabular}{l|cccccc}
    \hline    
    Model &IR$\uparrow$  &RR (@0.1m)$\uparrow$  &MRE$\downarrow$  &MTE$\downarrow$  \\ \hline
    (a.1) Angle-I2P (\textit{full}) &\textbf{49.5}  &\textbf{78.5}  &3.237  &\textbf{0.075} \\
    (a.2) Angle-I2P w/o scale alignment &48.7  &77.5  &\textbf{3.153}  &\textbf{0.075}\\ \hline
    (b.1) Angle-I2P (\textit{full}) &\textbf{49.5}  &\textbf{78.5}  &3.237  &\textbf{0.075} \\
    (b.2) Angle-I2P w/ distance-based spatial consistency &48.5  &76.8  &\textbf{3.216}  &0.077\\ \hline
    (c.1) Angle-I2P (\textit{full}) &\textbf{49.5}  &\textbf{78.5}  &\textbf{3.237}  &\textbf{0.075} \\
    (c.2) Angle-I2P w/o global-to-local cross-attention &49.0  &77.8  &3.261  &0.078\\ 
    (c.3) Angle-I2P w/o reweight weights &45.0  &76.9  &3.516  &0.083 \\ \hline
    (d.1) Angle-I2P w/ $\tau=0.2$ (ours) &49.5  &\textbf{78.5}  &\textbf{3.237}  &\textbf{0.075} \\
    (d.2) Angle-I2P w/ $\tau=0.4$ &54.7  &76.7  &3.265  &0.081  \\
    (d.3) Angle-I2P w/ $\tau=0.5$ &\textbf{58.0}  &75.8  &3.305  &0.078 \\ \hline
    \end{tabular}
    \label{Table 4: Ablation studies.}
\end{table*}

\noindent \textbf{Datasets.} RGBD Scenes V2 \cite{lai2014unsupervised} contains 14 different indoor scenes. In this set of experiments, we evaluate models trained on Scenes 1-9. The trained model is then tested on Scenes 11-14 to get the initial correspondences, then we get 1748 training pairs, 497 testing pairs and 236 valid pairs. Note that the test scenes is unseen, we leverage this dataset to validate the performance of our outlier removal method in eliminating mismatches for cross-scene registration.  

\noindent \textbf{Evaluation Results.} The results are shown in Table \ref{Table 2: Evaluation Results on Cross-scene Registration}. Due to the low number of initial correspondences $\mathcal{C}$ and low inlier ratio, methods like GoMatch \cite{zhou2022geometry} (which do not incorporate geometric constraints) failed to effectively filter outliers. Similarly, the performance of GraphI2P \cite{bie2025graphi2p} in outlier removal was suboptimal, as it lacks hierarchical processing integrating global-to-local information. Our method outperforms the second best GraphI2P by 5.9 percent on \textit{Inlier Ratio} and 0.9 percent on \textit{Registration Recall}. We also visualize the results in Fig. \ref{Figure 4: Visualization of RGBDScenesV2 Datasets.}.

\subsection{Evaluation Results on Self-Collected Datasets}

\noindent \textbf{Datasets.} To effectively validate the performance of our model in real-world scenarios, we captured data from eight distinct indoor scenes in a laboratory setting using an Intel Real Sense RGB-D camera. We evaluated the pre-trained 2D3D-MATR \cite{li20232d3d} model on the 7Scenes dataset across all eight scenes, obtaining corresponding results for both the training and testing sets. 

\noindent \textbf{Evaluation Results.} The results are shown in Table \ref{Table 3: Evaluation Results on Self-collected Datasets.}. On \textit{Inlier Ratio}. our method outperforms the second best GoMatch by 2.9 percent. And on the most important metric, \textit{Registration Recall}, our method surpass GraphI2P by 6.6 percent. These results demonstrate the enhanced generalization capability and potential for practical applicability of our model.   

\subsection{Ablation Studies}
\label{section 4.5 ablation studies}

\noindent \textbf{Effectiveness of scale alignment.} We remove the scale alignment module and denoted it as "\textit{Angle-I2P w/o scale alignment}". The results are shown in Table \ref{Table 4: Ablation studies.}(a). Without scale alignment, the estimated point cloud coordinates derived from the initial features contain errors. These errors propagate into the extracted features, consequently hindering the precise elimination of outliers within the image-point cloud correspondences. This ultimately leads to a decline in both the \textit{Inlier Ratio} and the \textit{Registration Recall}.

\noindent \textbf{Effectiveness of angle-based spatial consistency.} To effectively validate the role of angle-based spatial consistency, we use the original distance-based spatial consistency and denoted it as “\textit{Angle-I2P w/ distance-based spatial consistency}”. The results are shown in Table \ref{Table 4: Ablation studies.}(b). Angle-based spatial consistency inherently eliminates the influence of scale errors, whereas distance-based spatial consistency imposes stricter requirements on coordinate accuracy. Removing the angle-based component allows scale errors to compromise the precision of this geometric constraint, consequently adversely affecting metrics such as \textit{Inlier Ratio}, \textit{Mean Rotation Error}, and \textit{Mean Translation Error}.

\noindent \textbf{Effectiveness of global-to-local cross-attention.} To effectively validate the role of global-to-local cross-attention, we removed this mechanism from our model and denoted it as “\textit{Angle-I2P w/o global-to-local cross-attention}”. The results are shown in Table \ref{Table 4: Ablation studies.}(c). In rigid point cloud registration, global information should also be considered. Neglecting global context would result in locally similar confidence scores for correspondences, causing filtered inliers to cluster excessively. This concentration ultimately leads to reduced accuracy in the estimated pose.

To validate the role of spatial consistency weighting in the transformer architecture, we remove the consistency weights and denote this variant as \textit{“Angle-I2P w/o reweight weights”}. The results indicate that without the guidance of consistency weighting, the model lacks a robust basis for distinguishing between inliers and outliers. The lackness of the weights leads to a 4.5 pp drop in the inlier ratio of the filtered matches compared to the full Angle-I2P model.

\noindent \textbf{Effectiveness of inliers/outliers threshold.} Finally, we study the effectiveness of threshold $\tau$ in Table \ref{Table 4: Ablation studies.}(d). As the threshold increases, the inlier ratio rises while the registration accuracy decreases. This occurs because a higher threshold filters out more points. Additionally, neighboring points often exhibit similar spatial features, leading to comparable inlier scores. Consequently, the filtered points become more clustered, which can cause pose estimation algorithms to converge to incorrect solutions. Therefore, we adopt a balanced approach by setting the threshold to 0.2, which jointly improves both the \textit{Inlier Ratio} and \textit{Registration Recall}.

\subsection{Limitation and Future Work} 
In this paper, the analysis of scale errors introduced by the monocular depth estimator remains relatively coarse, and the global scale as well as bias have not been adequately corrected to align the estimated point cloud with the actual one. Additionally, due to the coarse-to-fine matching strategy employed by 2D3D-MATR \cite{li20232d3d}, the resulting matching pairs tend to form clusters. After filtering, these matches exhibit an uneven distribution, which leads to an improvement in IR (\textit{Inlier Ratio}) but not a significant gain in RR (\textit{Registration Recall}). These issues also require further investigation in future work.

\section{Conclusion}
In this paper, to effectively address the issue that existing image–point cloud registration methods still yield many outliers in the initial image–point cloud matches, we propose an end-to-end trainable framework for image–point cloud mismatch removal. This method can effectively eliminate mismatches in cross-modal matching and improve registration performance. We observe that in existing feature learning-based image-to-point cloud correspondences, many correspondences violate geometric consistency due to issues such as scale ambiguity. To address this, we design a scale-invariant cross-modal geometric constraint, combined with a global-to-local attention mechanism to effectively filter outliers. Experimental results across few-shot and cross-scenario settings demonstrate the superiority of our approach, showcasing enhanced outlier rejection performance and generalization
capability. So we believe it has the potential to contribute to advancements in image–point cloud mismatch removal.

\section*{Acknowledgement}
This work was supported in part by the National Key Research and Development Program of China under Grant 2024YFC3015302, in part by the National Science Foundation of China under Grant 62502171, in part by the Key Research and Development Program of Hubei Province under Grant 2024BAB021 and Grant 2024BAB036. 

\bibliography{mybibfile}

\end{document}